\documentclass[a4paper]{article}
\usepackage{anyfontsize}
\usepackage{INTERSPEECH2021}
\usepackage[table]{xcolor}
\usepackage{booktabs}
\usepackage{graphicx}
\usepackage[noadjust]{cite}
    
\usepackage{hyperref}
\usepackage{caption}
\usepackage{amsmath}
\usepackage{graphicx} % \scalebox
\usepackage{environ}
\usepackage{bm}
\usepackage{float}
\title{  FANS: Fusing ASR and NLU for on-device  SLU }
\name{Martin Radfar, Athanasios Mouchtaris,  Siegfried Kunzmann, and Ariya Rastrow}
\address{Alexa Machine Learning, Amazon, USA }
\email{\{radfarmr,mouchta,kunzman,arastrow\}@amazon.com}

\begin{document}

\maketitle
\begin{abstract}

Spoken language understanding (SLU) systems translate  voice input commands to   semantics which are encoded as  an intent and pairs of slot tags and values.  Most current SLU systems deploy a cascade of two  neural models where the first one  maps the input audio to a transcript (ASR) and the second  predicts the intent and slots from the transcript (NLU). In this paper, we introduce FANS, a new end-to-end  SLU model that fuses an ASR audio encoder to a multi-task NLU decoder to infer  the  intent, slot tags, and slot values directly from a given input audio, obviating the need for transcription. FANS consists of a shared audio  encoder and three decoders, two of which are seq-to-seq decoders that predict non null slot tags and slot values in parallel and in an auto-regressive manner. FANS neural encoder and decoders architectures are flexible which allows us to leverage different combinations of LSTM, self-attention, and  attenders. Our experiments show compared to the state-of-the-art  end-to-end SLU models, FANS reduces ICER and IRER errors relatively by 30\%  and 7\%, respectively, when tested on an in-house SLU dataset and by  0.86\% and 2\%  absolute when tested on a public SLU dataset.
\end{abstract}


\section{Introduction}

Spoken language understanding (SLU) systems  translate  the meaning of  a spoken utterance  to a format that is readable by machine \cite{wang2005spoken,TurDeMori2011,Bhargava2013EasyCI}. Today's commercial neural SLU systems deploy a cascade of ASR and NLU to infer semantic information from audio \cite{DBLP:journals/ftir/LarsonJ12}. In these models,  after a spoken utterance is transcribed by  an ASR module,  an NLU system decodes the desired NLU output from the transcript. This disjoint SLU framework has several drawbacks: First, ASR and NLU are optimized independently and NLU performance highly depends on ASR accuracy.  Second, many ASR errors occurred in  words that  have  null slot tags and, therefore, have no impact on NLU performance. Accordingly, training ASR to get a full transcript is sub-optimal for NLU. Third, disjoint SLU models restrict the model compression and sharing parameters end-to-end, a requirement for on-device SLU where  computational resource is limited. 

In this paper, we go beyond disjoint SLU paradigm  and introduce an end-to-end SLU system which is optimized to maximize the NLU accuracy directly from audio.  
The outputs of  an SLU system are often: an intent and two sequences that contain slot tags and values. In recent years, several end-to-end SLU approaches have beed proposed \cite{lugosch2019speech,renkens2018capsule,chen2018spoken,tomashenko2019recent,vu2016bi,liu2017topic,chen2019transfer,haghani2018audio,huang2019adapting,tomashenko2020dialogue,dinarelli2020data,wang2020large,lai2020semisupervised,1910.11559,rao2020speech,martinInterspeech2020,morais2020end,qian2017exploring,chung2021splat}  which can be classified into two broad categories: The models in the first category approach  SLU  as a multi-label classification problem aiming mostly to predict the  intent of a spoken utterance \cite{lugosch2019speech,vu2016bi,serdyuk2018towards,liu2017topic,tian2020improving}.  In \cite{serdyuk2018towards}, stacked layers of biGRU  along with  MLP are used to classify intent directly from audio files. In \cite{lugosch2019speech},  a SincNet layer \cite{Ravanelli2018}  is augmented to the architecture proposed in \cite{serdyuk2018towards} and leverage a pre-training strategy to further improve accuracy. 
 In addition to RNNs, convolutional neural networks (CNNs) also have been used alone or joint with RNNs for speech-to-intent classification \cite{liu2017topic,tian2020improving}.
In \cite{tian2020improving},    a cascade of CNN and GRU layers is used with different training strategy called Reptile, to infer the intents.  
  In \cite{9053281}, a multi-modal model was proposed in which a pre-trained text embedding incorporated as an auxiliary model along with  a biLSTM audio encoder to predict intent. 
 
The second category of SLU models infer not only intent but also slot tags and slot values. There have been different proposals, though less than the first category, to achieve this goal \cite{lee2015spoken,haghani2018audio,lai2020semisupervised,1910.11559,rao2020speech,martinInterspeech2020,morais2020end,qian2017exploring,chuang2020speechbert} . In  \cite{1910.11559}, an SLU model is proposed that  leverages  text BERT pre-training that is jointly optimized with ASR.   In \cite{rao2020speech}, a jointly trained  ASR and NLU is proposed by introducing an interface to couple ASR and NLU neural models. In this model, ASR and NLU are first pre-trained and then fine-tuned on the SLU task.  In  \cite{kuo2020end}, an SLU model is proposed that adapts the pre-trained CTC- and attention- based ASR modules to directly predict NLU output.
In \cite{martinInterspeech2020}, a transformer-based SLU is proposed  which is shown  to be a competitive E2E SLU models but this architecture  assumes the slot values are associated to a fixed, ordered, pre-defined slot tags. In \cite{morais2020end}, another transformer based SLU model  was proposed that leverages self-supervised pre-trained acoustic features, pre-trained model initialization that allows to use large training data and fine tuned on SLU data.  In \cite{haghani2018audio}, multi-task learning is leveraged for the SLU task. The authors proposed different model architectures to decode  ASR and NLU outputs  alone or together. In this multi-task learning scheme, one task is to decode the transcript and the other task is to decode serialized semantics (slot values and slot tags). They also proposed a direct model to decode the serialized slot tag and slot values.

In this paper, we propose, FANS, a new end-to-end neural  architecture that  fuses  the ASR  encoder directly to  the NLU decoder. FANS infers intent, slot tags, and slot value directly from audio with no dependency on transcription. In contrast to  \cite{haghani2018audio}, FANS avoids serialized semantics by deploying two parallel decoders  to infer only non null slot tags and their corresponding slot values. In addition, FANS does not use transcription which is used in  \cite{haghani2018audio} as an auxiliary task to help predict semantic. Using  attention mechanism, FANS learns to attend to  audio segments that are  solely relevant to NLU. FANS uses predicted intent as a prior to help better slot tag prediction. FANS model architecture leverages both LSTM and transformer in the encoder and decoder with different attention mechanisms to get best of both worlds. FANS learns parameters that are optimized for the NLU output rather than ASR so that makes FANS more accurate, compact, and a promising candidate for on-device SLU. We evaluate the performance of FANS on both public and in-house SLU datasets. Our experiments show a significant performance improvement when FANS is compared with  the state-of-the-art-models.

 The rest of this paper is organized as follows:   In Section 2,  we describe and formalize FANS architecture with its variants. In Section 3,  we demonstrate the superiority of FANS over a competitive model using both in-house and public datasets. Finally, in Section 4, we draw the conclusions.

\begin{figure}[h]
  \centering
  \includegraphics[width=.8\linewidth]{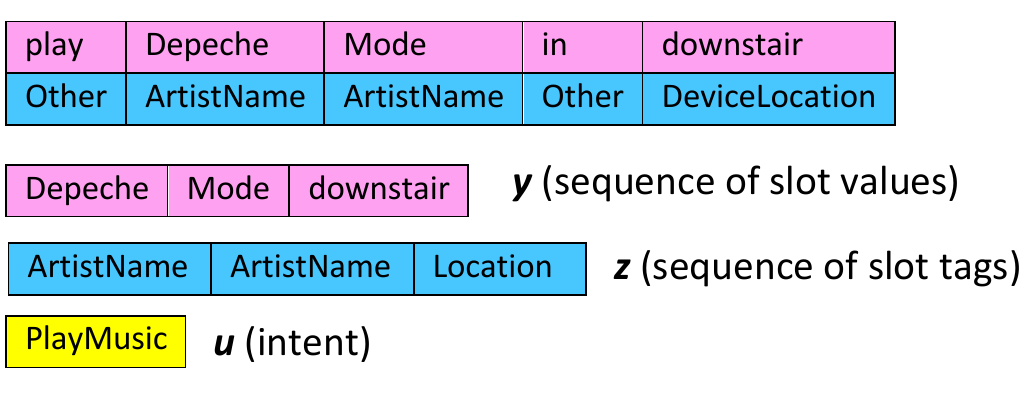}
    \caption{ The FANS output target consists of an intent  and two sequences that contain pair-wise match slot tag and slot values. We discard all slot values associated with slot labeled as Other--- note that we also use the term Null interchangeably in the text.}
  \label{fig:lossfunction}
\end{figure}

\section{FANS Architecture}
\subsection{FANS input and output structures}
The input audio signal is transferred to a time-frequency  space represented by  $\bm{x}=[x_1,\ldots, x_i,\ldots,x_T]^\mathsf{T}$  where $x_i \in \mathcal{R}^D$ , $T$ and $D$ denote the number of frames and the dimension of the time-frequency space, respectively;  also $\mathsf{T}$ denotes the transpose operation. Each audio signal, $\bm{x}$, has a transcript and a slot tag annotation represented by  $\hat{\bm{y}}=[\hat{y_1},\ldots, \hat{y_j},\ldots,\hat{y_L}]^\mathsf{T}$ and  $\hat{\bm{z}}=[\hat{z_1},\ldots, \hat{z_j},\ldots,\hat{z_L}]^\mathsf{T}$, respectively, where $L$ is the number of words in the transcript and
 $z_j \in  \{\text{vocab}(\mathcal{Z}) \cup Null \}$. In other words, each word in the transcript belongs either to a slot tag from a slot tag dictionary or it is labelled as $Null$ (or Other). An example is demonstrated in Figure~\ref{fig:lossfunction} in which the transcript is $\hat{\bm{y}} =[play, Depeche, Mode, in, Downstairs]^\mathsf{T}$ and $\hat{\bm{z}} =[Null, ArtistName, ArtistName, Null , DeviceLocation]^\mathsf{T}$. For the SLU task, we are only interested in inferring non $Null$ slot tags and words in the transcripts associated with non $Null$ slot tags---we call them slot values.  Accordingly, we exclude all $Null$ tags and words associated with the $Null$ tags to generate two new sequences: $\bm{y} =[ Depeche,Mode,downstairs ]^\mathsf{T}$ and $\bar{\bm{z}} =[ ArtistName,ArtistName, DeviceLocation]^\mathsf{T}$. As such, the target output would be  the slot tag $\bm{y}=[y_1,\ldots, y_j,\ldots,y_M]^\mathsf{T}$, and the slot value $\bar{\bm{z}}=[\bar{z_1},\ldots, \bar{z_j},\ldots,\bar{z_M}]^\mathsf{T}$, where $M$ is the length of the sequence after discarding $Null$ tags in the tag sequence and corresponding slot values. In addition to $\bm{y}$ and $\bar{\bm{z}}$, there is an intent, $u\in \text{vocab}(\mathcal{U})$, corresponding  to each $\bm{x}$. We insert $u$ as the first element into $\bar{\bm{z}}$ to get  $ \bm{z} =[u;\bar{\bm{z}}]^\mathsf{T}$.  Hence, we generate two sequences $\bm{y}$ and $\bm{z}$ which comprise the output of FANS and  together tuple $(\bm{x},\bm{y},\bm{z})$ forms the input and outputs of FANS, where $\bm{x}$ is a vector of audio features, $ \bm{y}$ a vector of slot values, and $\bm{z}$ is a vector of slot tag plus intent. 
 
 Different from our output structure, in  \cite{haghani2018audio}, the slot tag and slot values are serialized---what the authors called serialized semantics; for our example,  the serialized semantics given by  $ \hat{\bm{s}} =[ ArtistName,Depeche,ArtistName,Mode,\\ DeviceLocation,Downstairs]^\mathsf{T}$

 \subsection{FANS model} 
\begin{figure}[t]
  \centering
  \includegraphics[width=.8\linewidth]{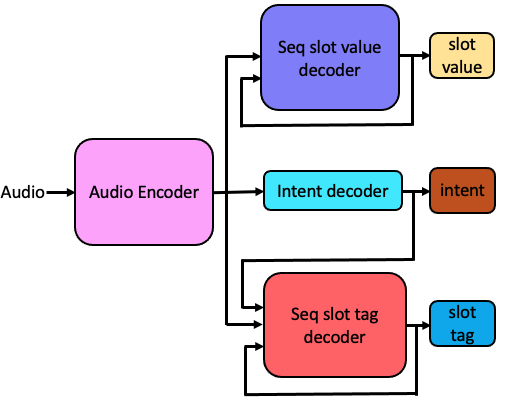}
  \caption{ FANS architecture. The model consists of a shared audio encoder, three decoders, and an attender .}
  \label{fig:fans}
\end{figure}

FANS consists of a shared audio encoder and two sequence decoders: slot tag and slot value decoders and one intent decoder whose parameters are shared with the slot tag decoder (Figure~\ref{fig:fans}). The neural structure of the encoder and the decoders can be any types of known neural architectures but here we use  different combinations of LSTM and self-attention---self-attention is technically  the self-attention mechanism used in Transformers \cite{vaswani2017attention} .  In addition, FANS deploys an attender that helps the decoders to attend to the encoder for better prediction.  We use both  additive attention \cite{bahdanau2014neural} and cross-attention  \cite{vaswani2017attention}.

The shared audio encoder  transforms $\bm{x}$ to the  higher level feature representations $\bm{h}^{\text{enc}}= [h_1,\ldots,h_i,\ldots,h_T]^\mathsf{T}$  where $h_i \in \mathcal{R}^K$, and $K$ is the dimension of the  encoder output feature space.    Given the input vectors, previous decoded outputs, and the model parameters $\Omega=\{\Omega^{\text{enc}},\Omega^{\text{dec-tag}},\Omega^{\text{dec-value}},\Omega^{\text{dec-intent}}\}$, FANS infers the slot values, slot tags, and intent by maximizing the following posterior probabilities: 
 \begin{equation}
 \begin{aligned}
\overset{*}{y_i}& = \underset{\text{vocab}(\mathcal{Y})}{\text{argmax}} \,\, p(y_i |\bm{x},f_{\text{att}}(\bm{h}^{\text{enc}},{\bm{y}}_{-i}),\Omega^{\text{enc}},\Omega^{\text{dec-tag}}),\\
\overset{*}{z_i} &= \underset{\text{vocab}(\mathcal{Z})}{\text{argmax}} \,\, p(z_i |\bm{x},f_{\text{att}}(\bm{h}^{\text{enc}},{\bm{z}}_{-i},u),\Omega^{enc},\Omega^{\text{dec-value}}),\\
 \overset{*}{u} &= \underset{\text{vocab}(\mathcal{U})}{\text{argmax}} \,\, p(u |\bm{x},\bm{h}^{\text{enc}},\Omega^{enc},\Omega^{\text{dec-intent}}),\\
 \end{aligned}
 \end{equation}
 where the subscript $-i$ represents all previous decoded tokens before the $i$th token. Also $f_{\text{att}}(\cdot,\cdot)$ denotes the attender function which is either the cross attention of the form:
 \begin{equation}
 f_{CA}(\bm{h}^{\text{enc}},{\bm{q}}_{-i}) = \text{softmax}\Bigl( \frac{{\bm{w}_q\bm{q}}_{-i}{\bm{h}^{\text{enc}}}^\mathsf{T}{\bm{w}_k}^\mathsf{T}}{\sqrt{K'}}\Bigr)\bm{w}_v\bm{h}^{\text{enc}}
 \end{equation}
where ${\bm{q}}_{-i}= {\bm{y}}_{-i}$ for the slot value decoder and ${\bm{q}}_{-i}= {\bm{z}}_{-i}$ for the slot tag decoder; also, $\bm{w}_q$, $\bm{w}_k$, and $\bm{w}_v$ are learnable query, key, and value matrices that transform the input vector into a  $K'$-dimensional space. The additive attention  given by 
\begin{equation}
 f_{AA}(\bm{h}^{\text{enc}},{\bm{q}}_{i-1}) =  \sum_{j=1}^{T} \text{softmax}\bigl(g({\bm{q}}_{i-1},\bm{h}^{\text{enc}}_{j})\bigr) \bm{h}^{\text{enc}}_{j}
 \end{equation}
 where ${\bm{q}}_{i-1}= {\bm{y}}_{i-1}$, the hidden state just before emitting ${\bm{y}}_{i}$ for  the slot value decoder and ${\bm{q}}_{i-1}= {\bm{z}}_{i-1}$, the hidden state just before emitting ${\bm{z}}_{i}$ for  the slot tag decoder. Also,  $g(\cdot)$ is a feedforward neural network. Given $N$ training samples drawn from distribution $p(\bm{x},\bm{y},\bm{z})$, the model parameters $\Omega$ are learned by maximizing the likelihood function
 \begin{equation}
 \begin{aligned}
\overset{*}{\Omega} = \underset{\Omega}{\text{argmax}}   \underset{\Lambda}{\sum} \Bigl( \log p(\mathcal{Y }|\mathcal{X},\Omega^{\text{enc}},\Omega^{\text{dec-tag}})&\\
 +\lambda_1 \log p(\mathcal{Z} |\mathcal{X},\mathcal{U},\Omega^{enc},\Omega^{\text{dec-value}})&\\
+ \lambda_2 \log p(\mathcal{U} |\mathcal{X},\Omega^{enc},\Omega^{\text{dec-intent}})\Bigr)\\
 \end{aligned}
 \end{equation}
where $\Lambda=(\mathcal{X},\mathcal{Y},\mathcal{Z},\mathcal{U})$ is  the set of the training data and $\lambda_1$ and $\lambda_2$ are two tuning parameters that can control the contribution of each likelihood function during the training process.

\begin{figure}[t]
  \centering
  \includegraphics[width=.9\linewidth]{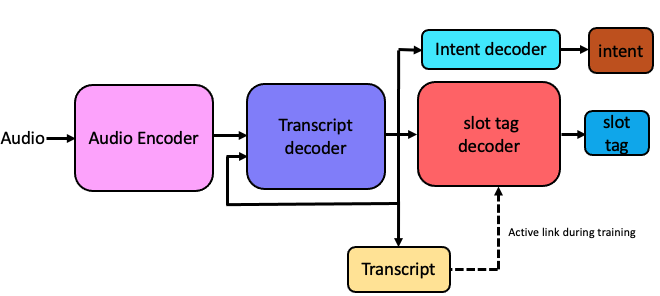}
  \caption{ A high-level block-diagram of cascade ASR-NLU SLU models.}
  \label{fig:slutask}
\end{figure}

 \section{Experiments}
 \subsection{Datasets and evaluation metrics}
 We use  two SLU datasets:  a public and  an  in-house. Compared to ASR and NLU datasets that contain thousands of  hours of transcribed audio files and numerous text, current popular public SLU  datasets have limited audio files. Three popular ones are FSC,  containing 30,000 utterances \cite{lugosch2019speech}, SNIP containing ~14,000 utterances \cite{coucke2018snips}, and ATIS, containing ~4,500 utterances \cite{price1990evaluation} which makes training SLU models end-to-end very challenging.   We opted for FSC  \cite{lugosch2019speech}  to be able to leverage end-to-end training. This dataset contains 11 intents, 3 slot tags, and slot values belong to a vocabulary of size 100. Because FSC is not annotated for intent, we consider the action labels as intents and the object and location labels as slot tags.  The SNIP and ATIS datasets are too small to be used for E2E training so these datasets usually are used as fine tuning datasets when models are pre-trained with large ASR and NLU corpa separately.
 
 We also used a subset of  500 hours of de-identified in-house SLU data in which  the data were split into  train (80\%), evaluation (10\%) and test (10\%) sets. Our SLU dataset contains 57 intents and 175 slot tags and the slot values belong to a vocabulary of size $\sim$100,000. We used 64-dimensional log short time Fourier transform  vectors obtained by segmenting the utterances with a Hamming window of the length 25 ms and frame rate of 10 ms. The 3 frames are stacked with a skip rate of 3, resulting  in 192-dimensional input features; these features are normalized using the global mean and variance normalization.
 
We evaluate the SLU models using  two well-known metrics:  intent classification error rate (ICER)  and interpretation error rate (IRER). These metrics are defined as follows. Let $I$ denote the number of utterances with miss-recognized intent. Let $E$  denote the number of utterances  that contains  any of miss-recognized intent,  slot tags, or slot values; e.g., if we have an utterance with one intent, 10 slot values and 10 slot tags and only one of these 21 (10+10+1) entities is miss-recognized, we count one error for the entire utterance.  Hence: $\text{ICER} =\frac{I}{N} $ and $\text{IRER} =\frac{E}{N}$
%\begin{equation}
%\text{ICER} =\frac{I}{N} \,\,\, \text {and}\,\,\, \text{IRER} =\frac{E}{N}
%\end{equation}
where $N$ is the total number of utterances. It should be noted for the in-house SLU data, we report  performance in terms of  relative increase in ICER (RI-ICER) and IRER (RI-IRER) against the best model, meaning that we set IRER and ICER to zero for the best model and compute the relative increase in ICER and IRER for other models when compared  to the best model.

\subsection{Model parameters}
We built three different versions of FANS by deploying LSTM and self-attention (SA) \cite{vaswani2017attention} neural structures in the encoder and decoders. We also used  cross-attention  (CA)\cite{vaswani2017attention}  or  additive attention (AA) \cite{bahdanau2014neural} as the attender.  We tried to keep all models within the  same size to make comparison not biased on the size of the model.  We compared FANS with the multi-task and direct A2I models proposed in \cite{haghani2018audio} . 

The three versions of FAN are as follows: 1- FAN-a: an LSTM based encoder, two LSTM based decoders, and an additive attention attender; 2- FAN-b: an LSTM based encoder, two self-attention based decoders, a cross-attention attender; 3- FAN-c: a self-attention based encoder and two self-attention based decoders, and  a cross-attention attender. In FAN-a, we used a $3\times 612$  LSTM in the encoder and two decoders.   In FAN-b, we used  a $4\times768$   LSTM in the encoder and 3 layers of self-attention layers in each decoder.  In FAN-c, we used 12 self-attention layers in the encoder, 5 self-attention layers in the slot value decoder, and 3 self-attention layers in the slot tag decoder. The input dimension (d-model) was set to 256. For all self-attention layers, we used 4 heads and a 2,048-hidden unit feedforward net is used. For the multi-task and direct A2I  in \cite{haghani2018audio},  the size of the encoder and decoder were set  to $3\times 612$  and $4\times 612$, respectively.  We trained these models using 500 hours of data in-house data.

For the FSC dataset, we built smaller models to prevent overfitting as the FSC dataset is very small and not complex. We observed models of size $\sim$ 3 millions deliver good results and don't suffer from overfitting. For FSC, the FANS models were built as follows:  In FAN-a, we used a $1\times 256$  LSTM in the encoder and two decoders.   In FAN-b, we used  a $2\times 232$   LSTM in the encode and 2 layers of self-attention layers in each decoder with  1 head and a 800-hidden unit feedforward net.  In FAN-c, we used 2 self-attention layers in the encoder, 1 self-attention layer in the slot value decoder, and 1 self-attention layer in the slot tag decoder with  4 heads and a 1024-hidden unit feedforward net.   The input dimension (d-model) for all these models was set to 128. For the multi-task and direct A2I  in \cite{haghani2018audio},  the size of the encoder and decoder were set  to  $1\times 256$  and $1\times 350$ LSTM, respectively.

The self-attention sub-space dimension (the column dimension of the query, key and value matrices) was set to 64.  The dropout rate and label smoothing were set to 0.1.  We used  Adam optimizer with $\beta_1$ = 0.9, $\beta_2$= 0.98, and $\epsilon$ = 1e-9.  The learning curve was chosen by setting the pre-defined factor $k$ and warm-up rate $w$ to 0.99 and 16,000, respectively. We used step size of 1000 and the model trained until no improvement observed.  For all experiments,  $\lambda_1 $ and $\lambda_2$ were set to 1. For the cascade ASR-NLU model we used 12/4  and 7/2 layers self-attention in the encoder and decoder of ASR for large/small models, respectively. For the NLU model, we used a transformer encoder with 2 layers that was trained on the full transcript and the NLU output including $Null$ tags . The large models were trained using four machines each of which has eight GPUs and the small models were trained on one GPU.

\iffalse
\begin{table}[t]
 \caption{Model parameters for architectures used in the experiments for both large ($\sim$ 30 M) and small ($\sim$ 3 M) models. Dec I : the slot value decoder; Dec II: slot tag decoder for all models but IV in which Dec II indicates the NLU decoder; hidd: number of hidden units in LSTM. d-ff: hidden size of feed-forward in transformer self-attention layer. d-m: dimension of the model.  }
   \label{tab:results}
 \centering
\rowcolors{1}{green!25}{blue!50}
\begin{tabular}{ *7l }    \toprule
\emph{Model} &Enc&Dec I& Dec II &Hidd &D-ff & D-m \\\midrule
I-a&3&3& 3&612&-& 256\\ 
\cline{2-7}
I-b&1&1&1&256&-& 256  \\ 
II-a & 4&4&-&612&-& 256\\
\cline{2-7}
II-b & 1&1&-&350&-& 256\\ 
III-a& 3&3&3&612&-& 256\\ 
\cline{2-7}
III-b&1&1&1&256&-& 256\\
VI-a & 4&3&3&768&2048& 256\\ 
\cline{2-7}
VI-b&2&2&2&232&800& 128\\
V-a & 12&5&3&-&2048& 256\\ 
\cline{2-7}
V-b&2&1&1&-&1024& 128\\
IV-a& 12&7&2&-&2048& 256\\ 
\cline{2-7}
IV-b&4&2&2&-&1024& 128\\\bottomrule
 \hline
\end{tabular}
\end{table}
\fi

\begin{table*}[t]
  \caption{Results in terms of relative increase in ICER  (RI-ICER) and IRER (RI-IRER) against the best model; The last column gives the model size in terms of the number of parameters in million; In this experiments we used  our in-house SLU dataset. SA: self-attention, CA: cross attention \cite{vaswani2017attention}   and AA: additive attention\cite{bahdanau2014neural} .}
   \label{tab:results}
 \centering
\rowcolors{1}{red!25}{yellow!50}
\begin{tabular}{ *7l }    \toprule
Model &Encoder&Decoder&Attender& \emph{RI-ICER (\%)} &  \emph{RI-IRER(\%)} &  Size (M) \\\midrule
A2I- multi-task \cite{haghani2018audio} &LSTM& LSTM &AA& 30 & 7&31\\ 
A2I-direct-model \cite{haghani2018audio} &LSTM&LSTM &AA & 39& 21&29\\ 
FANS-a & LSTM&LSTM&AA&25& 16 &31\\
FANS-b & LSTM&SA&CA&22 & 11&30\\ 
FANS-c & SA&SA&CA&0 & 0&30\\ 

Joint ASR NLU &SA&SA&CA& 47 & 35&30\\\bottomrule
 \hline
\end{tabular}
\end{table*}

\iffalse

\begin{table*}[t]
  \caption{ Results in terms of ICER and IRER, and  model size; In this experiments we used  our in-house SLU dataset.}
   \label{tab:results}
 \centering
\rowcolors{1}{red!25}{yellow!50}
\begin{tabular}{ *8l }    \toprule
Index&Model &Encoder&Decoder&Attender& \emph{ICER (\%)} &  \emph{IRER(\%)} &  Size (M) \\\midrule
I-a&A2I- multi-task \cite{haghani2018audio} &LSTM& LSTM &AA& 5.54 & 22.4&31.20\\ 
II-a&A2I-direct-model \cite{haghani2018audio} &LSTM&LSTM &AA & 6.42 & 26.51&29.20\\ 
III-a&FANS & LSTM&LSTM&AA&5.16 & 24.96 &31.20\\
IV-a&FANS & LSTM&SA&CA&5.01 & 23.49&30.10\\ 
V-a&FANS & SA&SA&CA&3.89 & 20.83&30.10\\ 

VI-a&Joint ASR NLU &SA&SA&CA& 7.41 & 35.93&30.3\\\bottomrule
 \hline
\end{tabular}
\end{table*}

\fi

\begin{table*}[t]
 \caption{Resutls in terms of ICER and IRER,  and model size; In these experiments, we used public FSC SLU dataset \cite{lugosch2019speech}. }
   \label{tab:results}
 \centering
\rowcolors{1}{blue!25}{yellow!50}
\begin{tabular}{ *7l }    \toprule
Model &Encoder&Decoder& Attender& \emph{ICER (\%)} &  \emph{IRER(\%)} &  Size (M) \\\midrule
A2I- multi-task \cite{haghani2018audio} &LSTM& LSTM& AA& 2.05 & 3.45&2.2\\ 
A2I-direct-model \cite{haghani2018audio} &LSTM&LSTM&AA  & 2.37 & 3.28&2.3\\ 
FANS-a & LSTM&LSTM&AA&2.14 & 3.03 &2.2\\
FANS-b & LSTM&SA&CA&0.71& 1.03&2.3\\ 
FANS-c& SA&SA&CA&2.03 & 3.9&2.3\\ 
Joint ASR NLU &SA&SA&CA& 1.56 & 14.49&2.31\\\bottomrule
 \hline
\end{tabular}
\end{table*}

\subsection{Results}
Table 1 and 2 show the RI-ICER/RI-IRER and ICER/IRER results which are obtained by using the in-house and FSC datasets, respectively. As shown in Table 1,  the best performance is pertained to FANS-c  which deploys self-attention both in the encoder and the decoders with a cross attention attender--- this model  is basically a transformer with three decoders. We observed  30 \%  and 7 \% relative reduction in ICER and IRER compared to the multi-task A2I  \cite{haghani2018audio}. We also observe averaging the ICER and IRER , all variants of FANS outperform three other models. Similar to what is reported in \cite{haghani2018audio}, we also observed the  A2I direct model underperforms compared to the A2I multi-task model suggesting that in the serialized semantic models having a transcript learner helps the NLU decoder. It is also seen that the joint ASR NLU model (Figure 3) exhibits worst performance confirming that the performance of these non E2E  SLU models highly depends on separately pre-trained ASR and NLU models.   A closer look at slot tag and slot values reveals the slot values prediction is more challenging as the number of slot values  are far exceeding the number of slot tags.  None of the models compared here are pre-trained.

We also compared FANS with other models when we used the public FSC SLU dataset as shown in Table 2. In order to avoid over-fitting, we reduced the size of the models to $\sim$ 3 millions. Using the FSC dataset, we observed  FANS-b outperforms all other models with 0.86 \% and 2.00 \% ICER  and IRER reduction, respectively.  We speculate two reasons as why FANS-b outperforms FANS-c on FSC: First, the utterances in FSC are very short so the attention mechanism in the encoder is not as effective as when we deal with longer utterances contained in the larger datasets. Second, FSC  is not semantically complex and has very low entropy with repetitive utterances and small number of  intent  and slot tags; as such, for this dataset, a smaller model is sufficient to provide satisfactory results.  We also observe the gap between ICER and IRER is smaller than large data set; this is however mainly because FSC has much smaller number of slot tags compared to the large dataset (3 vs 175).  We also observe IRER for the joint ASR NLU model is strikingly high; one explanation is that the 20 hours FSC is not enough to train the ASR model which requires to deliver transcript to the NLU modules with as low WER as possible. Overall, the results  given in Table 1 and 2 suggest that multi-task learners for slot tag and slot values is a better strategy than serializing them into one stream and using one learner.

\iffalse
As the results in Table 2 and 3 indicate the IRER for joint ASR NLU models is lower than the other models.  We conducted an experiment to  investigate impact of ASR error on the NLU outputs. Ideally, if we can get the transcript with no WER, the SLU system becomes an NLU where already the state of the art results reported using transformer based neural architectures (e.g. see \cite{wang2020encoding} and references therein). In our experiments, we injected synthetically generated error into the transcripts of the FSC data to resemble the WER introduced by  ASR in the  joint ASR-NLU SLU model and measured the slot tag errors. Figure~\ref{fig:WERNLU}) illustrates the slot tag error when WER increases from 0\% to 10\%.
We can see from the figure slot tag accuracy is very sensitive to ASR outputs even for FSC data for which we have three slot tags. This observation suggests that  the ASR-NLU SLU models require a reliable ASR module for which  increasing ASR model size and using more training data are two requirement which in turn introduce challenges in on-device SLU where we have limited computational resources and data.

\begin{figure}[t]
  \centering
  \includegraphics[width=.9\linewidth]{iFIG4.png}
  \caption{ The impact of WER on slot tag error rate for ASR +NLU SLU architecture. The NLU output is very sensitive to ASR output. }
  \label{fig:WERNLU}
\end{figure}
\fi
\section{Conclusions}
In this paper, we introduce FANS, a new neural architecture for end-to-end SLU. FANS fuses ASR and NLU modules to decode outputs that matters for NLU. FANS obviates the need for recognizing the transcript, allowing decoders  to attend to the parts of input acoustics that are relevant for inferring slot tags and slot values.  We showed that using different learners for slot tags and slot values leads to better performance compared to when all semantic information serialized.  Majority of words in ASR transcripts are  associated with  null slot tags; therefore what the NLU module expects from the audio encoder is a high level representation for words corresponds to non null slot tags. FANS architecture learns to only focus on these words.  We showed through experiments on both in-house data and public datasets that FANS outperforms competitive neural SLU models. The compact structure of FANS and its high accuracy make it a promising candidate for on-device SLU.
%\section{Acknowledgment}
%The authors would like to thank  Bjorn Hoffmeister for supporting this project and Rao Milind for FSC data preparation and pipeline.

\bibliographystyle{IEEEtran}

\bibliography{mybib}

\end{document}